\documentclass[10pt,twocolumn,letterpaper]{article}

\usepackage{3dv}
\usepackage{times}
\usepackage{epsfig}
\usepackage{graphicx}
\usepackage{amsmath}
\usepackage{amssymb}

\usepackage{booktabs}
\usepackage{mathtools}
\usepackage[normalem]{ulem}
\useunder{\uline}{\ul}{}
\usepackage[table,xcdraw]{xcolor}
\usepackage{multirow}
\usepackage{subcaption}
\usepackage{xr}

\newcommand{\new}[1]{\textcolor{black}{#1}}
\newcommand{\old}[1]{\textcolor{black}{}}

\usepackage[pagebackref=true,breaklinks=true,colorlinks,bookmarks=false]{hyperref}

\makeatletter
\renewcommand{\paragraph}{%
  \@startsection{paragraph}{4}%
  {\z@}{2.0ex \@plus 1ex \@minus .2ex}{-1em}%
  {\normalfont\normalsize\bfseries}%
}
\makeatother

\usepackage[capitalize]{cleveref}
\crefname{section}{Sec.}{Secs.}
\Crefname{section}{Section}{Sections}
\Crefname{table}{Table}{Tables}
\crefname{table}{Tab.}{Tabs.}

\threedvfinalcopy 

\def\threedvPaperID{66} 

\ifthreedvfinal\fi
\begin{document}

\title{Structure-Aware 3D VR Sketch to 3D Shape Retrieval}
\author{Ling Luo$^{1,2}$
\and 
Yulia Gryaditskaya$^{1,2,3}$
\and
Tao Xiang$^{1,2}$
\and
Yi-Zhe Song$^{1,2}$
\and\\
$^{1}$SketchX, CVSSP, University of Surrey, United Kingdom \\
$^{2}$iFlyTek-Surrey Joint Research Centre on Artificial Intelligence 
\\
$^{3}$Surrey Institute for People Centred AI and CVSSP, University
of Surrey, United Kingdom\\
}

\maketitle

\begin{abstract} 
We study the practical task of fine-grained 3D-VR-sketch-based 3D shape retrieval. This task is of particular interest as 2D sketches were shown to be effective queries for 2D images.
However, due to the domain gap, it remains hard to achieve strong performance in 3D shape retrieval from 2D sketches. 
Recent work demonstrated the advantage of 3D VR sketching on this task. 
In our work, we focus on the challenge caused by inherent inaccuracies in 3D VR sketches.
We observe that retrieval results obtained with a triplet loss with a fixed margin value, commonly used for retrieval tasks, contain many irrelevant shapes and often just one or few with a similar \textit{structure} to the query.
To mitigate this problem, we for the first time draw a connection between adaptive margin values and shape similarities.
\old{In particular, we propose to use a triplet loss with an adaptive margin value driven by a ``fitting gap'', defined as the similarity of two shapes under structure-preserving deformations.}
\new{In particular, we propose to use a triplet loss with an adaptive margin value driven by a `fitting gap', which is the similarity of two shapes under structure-preserving deformations.} 
We also conduct a user study which confirms that this fitting gap is indeed a suitable criterion to evaluate the structural similarity of shapes. 
\new{Furthermore, we introduce a dataset of 202 VR sketches for 202 3D shapes drawn from memory rather than from observation.}
The code and data are available at \url{https://github.com/Rowl1ng/Structure-Aware-VR-Sketch-Shape-Retrieval}.

\end{abstract}

\begin{figure*}
\includegraphics[width=\textwidth]{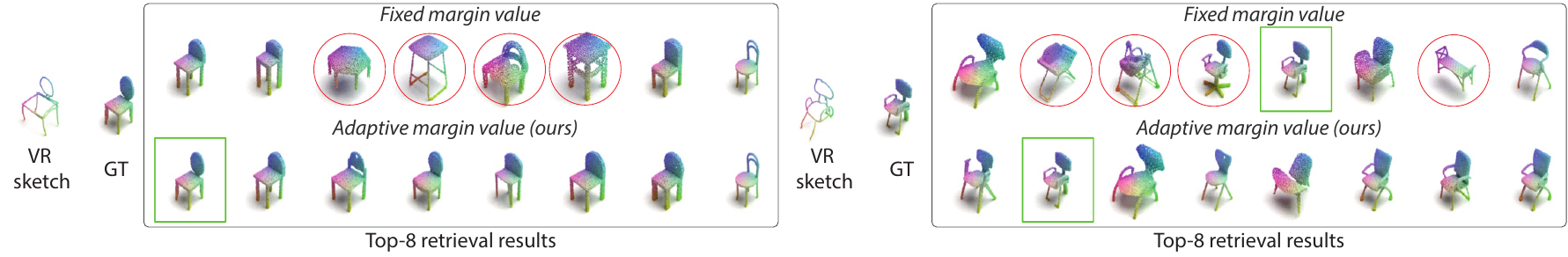}
  \vspace{-0.8cm}
  \caption{We propose an adaptive margin scheme within a triplet loss ubiquitously used for retrieval tasks. 
As shown in this figure, this strategy allows to obtain more `relevant' retrieval results that closer match the input sketch and the ground-truth 3D shape structure.
  The green square highlights the ground-truth 3D shape among the retrieval results, while the red circles mark erroneous retrieval results (dissimilar shapes). }
  \label{fig:teaser}
\end{figure*}

\section{Introduction}

\begin{figure*}[ht]
  \centering
    \includegraphics[width=\linewidth]{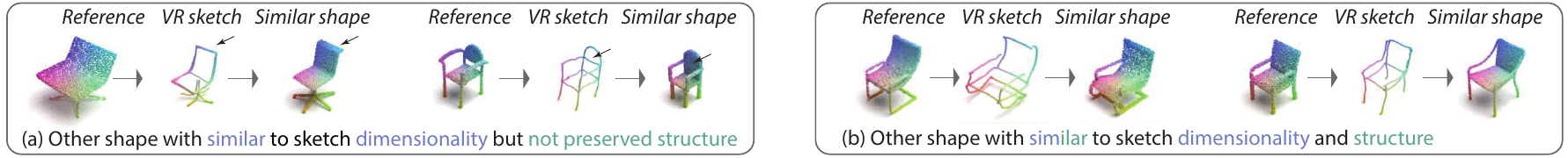}
    \vspace{-0.6cm}
   \caption{
   The VR sketches shown here were collected from the `reference' 3D shapes \cite{luo2021FineGrainedVR}. 
   The inherent distortions and sparsity of VR sketches pose challenges for the fine-grained 3D shape retrieval task.
   There may be a similar 3D shape that matches the overall sketch dimensions similar or better than the ground-truth but may have a worse (a) or better match (b) in structure and details than the ground-truth. 
   Therefore, we argue for a practical retrieval system that should be able to not only take the overall shape dimensions into account (a), but match the shapes with a similar structure (b). Likewise, the evaluation requires criteria that allow evaluating the structural similarity of the retrieved shapes (b). 
   }
   \label{fig:what_is_good_retrieval_match}
\end{figure*}

Progress in VR/AR technologies poses requirements on the availability of 3D content, thereby contributing to the active development of research on 3D shapes \cite{chen2020unsupervised,gadelha2020learning,yifan2020neural,cao2020comprehensive}. This is further accompanied by the growth of large 3D shapes collections. 
In our work, we study \emph{3D VR sketching}\new{, 3D sketching by means of a VR headset and VR controllers\footnote{Note that in general, a 3D sketch can also be created using Kinect or some 2D digital devices. As different sketching setups result in different quality of sketches and their distortions, we refer to sketches created with VR headsets/controllers as 3D VR sketches.},} in a minimal sketching interface as an intuitive tool for 3D shapes retrieval and exploration.

Sketch was proven to be an efficient media for 2D images retrieval \cite{yu2016shoe,deep-spatial-semantic}. 
However, despite recent attempts to address the domain gap between 3D shapes and 2D sketches \cite{qi2021Retrieval, liang2021uncertainty}, instance-level retrieval of 3D shapes from 2D sketches remains challenging at large. 
On the contrary, recent works \cite{luo2020towards,luo2021FineGrainedVR} demonstrate the \old{potential}\new{advantage} of 3D VR sketches\old{ as a convenient means} for 3D shapes retrieval. 
\new{Despite encouraging results, these models return many irrelevant shapes among the top retrieval results (\cref{fig:teaser}).}
\old{Despite encouraging results, the out-of-the-box retrieval model returns many irrelevant shapes among the top retrieval results (\cref{fig:teaser}).} 
This is largely due to the abstraction and distortion inherent in VR sketches.
In this work, we strive for a system that accounts for such inaccuracies during training and returns among the top retrieval results shape that are structurally similar to the query sketch.

Traditionally fine-grained retrieval is evaluated by ranking the retrieval results for each query according to a used distance metric and counting the percentage of instances that have ground-truth among the top-$k$ ranked results. 
As freehand sketches frequently contain distortions, top-$k$ accuracy does not allow to take into account the cases where the top retrieval results perfectly match the sketch but do not contain the ground-truth itself, as we show in \cref{fig:what_is_good_retrieval_match}. 
To measure structural similarities between 3D shapes, we adopt a \emph{fitting gap} \cite{uy2020deformation,uy2021joint} that evaluates how similar a given 3D shape is to another 3D shape under a structure-preserving deformation. 
In particular, we propose an \emph{adaptive margin scheme} for a triplet loss, commonly used for retrieval tasks  \cite{wang2014learning,schroff2015facenet,wen2016discriminative,grabner20183d}, to explicitly exploit this fitting gap.
We further leverage the recent progress in deep structure-preserving deformation methods \cite{yifan2020neural,jakab2021keypointdeformer} to compute the fitting gap on the fly during training. 
We demonstrate that our approach outperforms losses proposed in the context of deformation-aware retrieval \cite{uy2020deformation,uy2021joint} in the task of the structure-aware 3D shape retrieval from a 3D VR sketch.
Furthermore, for the first time, we leverage the fitting gap as an evaluation criterion for retrieval. To this end, we conduct a user study that shows that this criterion by far outperforms the Chamfer distance in measuring structural similarities, as judged by a human observer. 
Last but not least, we also show that this fitting gap-based criterion provides additional insights into retrieval performance.

In summary, our contributions include: (i) a novel formulation of a triplet loss with an adaptive margin value driven by the fitting gap between a ground-truth 3D shape and a given gallery shape, (ii) a perceptual study of 3D shapes distance measures on their effectiveness on evaluating structural similarities, (iii) adoption of the fitting gap as a distance metric to evaluate how well the retrieved results preserve structure and provides more fine-grained insights into retrieval performance, and (iv) a new test set of VR sketches drawn from memory.

\section{Related Work}
\label{sec:related}

\paragraph{Sketch Based Retrieval}
Our work is inspired by the success of fine-grained and category-level 2D-sketch-based image retrieval (SBIR) methods \cite{wang2015sketch,sangkloy2016sketchy,yu2016shoe,liu2017deep,collomosse2017sketching,deep-spatial-semantic,yu2021fine}. Similar to majority of these works we leverage the triplet loss, commonly used in the context of fine-grained and category-level retrieval \cite{wang2014learning,schroff2015facenet,wen2016discriminative,grabner20183d}. 

However, 2D-sketch-based 3D shape retrieval (SBSR) was mostly demonstrated in the context of \emph{category-level} retrieval: the pioneering hand-crafted feature based methods \cite{eitz2012sketch,yoon2010sketch,li2014shrec,li2016model} and the more recent deep learning based methods   \cite{xie2017learning,wang2015sketch,zhu2016learning,dai2017deep,he2018triplet,qi2018BMVC}. Qi \etal \cite{qi2021Retrieval} are the first to target fine-grained retrieval of 3D shapes from 2D sketches. 
They show that existing methods for Fine-Grained (FG) SBIR and category-level SBSR perform poorly on the task of FG-SBSR. 
Similarly to this work, we are interested in fine-grained retrieval of a 3D shape (retrieval of a particular shape instance), but consider a 3D VR sketch as a query.
In the supplemental, we show that our method outperforms \cite{qi2021Retrieval} results by $7.44$ points in top-1 retrieval accuracy on a similar training and test set sizes on a chair category. 

\paragraph{VR-Sketch-Based Shape Retrieval}
We are not the first to consider 3D-VR-SBSR, but as with 2D sketches, most of the existing methods for 3D-shape retrieval target category-level retrieval \cite{li20153d,li2015kinectsbr,li2016shrec,ye20163d,luo2020towards}. 
The earlier works \cite{li20153d,li2015kinectsbr,li2016shrec,ye20163d} relied on the dataset of 300 human-drawn 3D sketches of 30 classes, each with 10 sketches obtained with a Kinect-based virtual 3D drawing system (not available online anymore). 
Luo \etal \cite{luo2020towards} proposed a heuristics-based algorithm to generate synthetic 3D sketches from the reference 3D shapes, that they leveraged to train deep models. 
Giunchi \etal \cite{giunchi20183d,giunchi2021mixing} considered fine-grained retrieval of 3D shapes when a VR sketch is created on top of some reference 3D shape. 
In their case, the VR sketch comprises a set of dense ribbon-like strokes. Luo \etal \cite{luo2021FineGrainedVR} studied FG-3D-VR-SBSR from sparse quick sketches and introduced the first dataset comprising 1,497 3D VR sketch and 3D shape pairs of chairs. 
They performed benchmarking on common 3D shape retrieval methods and demonstrated encouraging results. 
In our work, we exploit their dataset and propose an adaptive margin scheme for a triplet loss, taking into account structural similarities between shapes.

\paragraph{Triplet loss with adaptive margin}
Triplet loss is commonly used for retrieval tasks  \cite{wang2014learning,schroff2015facenet,wen2016discriminative,grabner20183d}. It requires defining for each training set sample the sets of positive (`similar') and negative (`dissimilar') samples. 
It results in the latent space where `similar' instances are encoded closer than `dissimilar' ones.  It performs well in classification tasks, but is less suitable for the continuous ratings \cite{ha2021deep}, which is the case for our problem, as we want all top-$k$ retrieval results to be structurally similar to the query sketch. 
Variants of adaptive margin schemes were shown to be efficient for pose estimation \cite{zakharov20173d}, image retrieval \cite{zhao2019weakly,ha2021deep} and facial expression recognition \cite{tian2019outlier}.

\paragraph{Deformation-aware retrieval}
Our problem of retrieving a 3D shape that matches the structure of a 3D query sketch bears similarities with the problem of deformation-aware retrieval \cite{uy2020deformation,uy2021joint}. 
These works address the problem of retrieving a clean and complete 3D model from a noisy and/or partial 3D scan. 
They argue for the scenario where the goal is to retrieve the shape that after the deformation is the closest to the query. 
In this context, they define the \emph{fitting gap} as the difference between the deformed gallery model and the query. 
In our case, the \emph{fitting gap} is measured between gallery shapes and a ground-truth 3D shape of a query sketch. 
Our work is closest to \cite{uy2021joint}, where they exploit the fitting gap to ensure that the gallery shapes are embedded closer to the query shapes they can deform into. They leverage the regression loss, first proposed in \cite{uy2020deformation}.  
Instead, we propose a strategy for an adaptive margin value setting driven by a fitting gap. 
We show its superiority over the regression loss.
%
Further, Uy et al.~\cite{uy2021joint} jointly train for the deformation and retrieval in an alternating fashion, keeping one module fixed when optimizing the other and vice versa. 
Their deformation module and embedding space are optimized for deforming the retrieved shapes to the target shapes. In our work, the deformation module \emph{is trained first} and then used to guide the training to obtain the embedding space where shapes/sketches with similar structure are close to each other.
Our perceptual study indicates the validity of this measure for 3D shapes structural similarity, outperforming existing ones.
We aim at 3D VR sketch understanding and retrieval of a gallery shape that matches the query sketch best, prioritizing structural similarities.


%
%

%
%
%

\section{Method}
\label{sec:method}
Following \cite{luo2021FineGrainedVR}, we represent 3D shapes and 3D sketches as point clouds, and train the model via a Siamese training with a triplet loss \cite{wang2014learning,schroff2015facenet,wen2016discriminative,grabner20183d}. As a 3D sketch and shape encoder we exploit PointNet++\cite{qi2017pointnet++}, where the same set of weights is used to encode both modalities.

We denote the feature embedding of a sketch $S$ as $s \in \mathbb{R}^{512}$ (source) and a shape $T$ as $t \in \mathbb{R}^{512}$ (target). 
For a given batch of $N$ sketch-shape pairs, in \cite{luo2021FineGrainedVR} the triplet loss is defined as follows:
\begin{equation}
 L_T = {1}/{|N^n_S|}\sum_{t_S^{n} \in N^n_S} [d(s,t_S^{p})^2-d(s,t_S^{n})^2+m]_{+},
   \label{eq:TL}
\end{equation}
where $t_S^{p}$ is an embedding of the shape $T_S^p$ matching the considered sketch $S$. 
$N^n_S = \cup T_S^{n}$ is a set of negative shape examples, $T_S^{n}$, which are  
all shapes in a batch excluding $T_S^{p}$. 
$m$ is a margin, $[\cdot]_{+}$ denotes the clipping of negative values (a hinge loss), and $d$ is a Euclidean distance in the embedding space.
Feature vectors are $L_2$ normalized.
The triplet loss ensures that the source sketch and its matching shape are closer in the feature space than the source sketch and all other 3D shapes. 

We propose an adaptive margin that determines how far to push the embeddings of a sketch $s$ and its matching shape from the embeddings of shapes $t_S^{n}$ in the negative set of shapes based on how similar the 3D shapes $T_S^{p}$ (sketch matching shape) and $T_S^{n}$ (some shape from the set of negative shapes) are.
In particular, let $\delta(\cdot,\cdot)$ be some distance between two 3D shapes. 
Then, for each shape, we compute distances to each other shape. Let $M_S(T_S^{p}, N^n_S)$ be the maximum distance from the shape $T_S^{p}$ to shapes in $N^n_S$. 
We compute the margin value for each pair of shapes $T_S^{p}$ and $T_S^{n}$ as follows:
\begin{equation}
\label{eq:margin}
 m \coloneqq m(T_S^{p},T_S^{n}) = \alpha + (\beta - \alpha)\frac{\delta(T_S^{p},T_S^{n})}{M_S(T_S^{p},N^n_S)},
\end{equation}
where $\alpha$ and $\beta$ are method hyper-parameters.
This mapping has nice properties: in the unlikely case that all the shapes in the batch are similar, this loss will try to push them far apart, still satisfying fine-grained retrieval requirements. 
In the more likely case and on condition of a sufficiently large batch size, $M_S(T_S^{p}, N^n_S)$ will be similar between batches, ensuring consistent behavior of the mapping driven only by $\delta(T_S^{p},T_S^{n})$. 


\subsection{Choice of the distance $\delta$ in 3D space}
\label{sec:deformer}
One obvious choice of the distance between two 3D shapes is the Chamfer distance \cite{barrow1977parametric}, which is by far the most commonly used distance for training and evaluating similarities between two 3D shapes \cite{fan2017point}.
In this work, we consider additional criteria to compare 3D shapes which prioritize structural similarities.

First, we define an \emph{asymmetric fitting gap} as the distance between the sketch matching shape $T_S^{p}$ and the deformed target/gallery shape $T_S^{n}$ to the $T_S^{p}$:
\begin{equation}
\label{eq:fiting_gap}
\delta^A_{CD}(T_S^{p},T_S^{n}) \coloneqq CD(\mathcal{D}(T_S^{n};T_S^{p}),T_S^{p}), 
\end{equation}
where $\mathcal{D}(T_S^{n};T_S^{p})$ is a deformation operator that deforms $T_S^{n}$ to $T_S^{p}$, and $CD$ denotes the Chamfer distance. We refer to this definition of $\delta^A_{CD}(\cdot,\cdot)$ as the \emph{asymmetric fitting gap} since we only deform the target shapes to the source ground-truth (sketch matching) shape.
\new{The intuition behind the asymmetric fitting gap follows the findings by Uy et al.~\cite{uy2020deformation} that while one shape can be easily deformed into another, it might not be true in the opposite direction. The direction of the deformation is driven by the observation that the structure of the retrieval results is more likely to be what the user had in mind if the retrieval results can be deformed to the target shapes. }

Then, we define a \emph{symmetric fitting gap} $\delta^S_{CD}(T_S^{p},T_S^{n})$ as:
\begin{equation}
\label{eq:fiting_gap_s}
1/2\left(CD(\mathcal{D}(T_S^{n};T_S^{p}),T_S^{p}) +  CD(T_S^{n}, \mathcal{D}(T_S^{p};T_S^{n})) \right).
\end{equation}
The rational behind the symmetric fitting gap is that if the deformation can be accurately performed in both directions, then the shapes are more likely to have the similar structure. 
However, the risk is that the results for some shapes may be biased due to the efficiency of the deformation of the source into the target shapes. 

As the deformation operator, we leverage a recent neural deformation operator \cite{yifan2020neural} that allows efficient computation of the margin value during training\new{, and was shown to work well on diverse classes of shapes.}
We train the deformation operator on the 3D shapes from the chair class of the ShapeNet dataset. 
We visualize the deformation results in supplemental materials.




\begin{figure}[t]
  \centering
\includegraphics[width=1.0\linewidth]{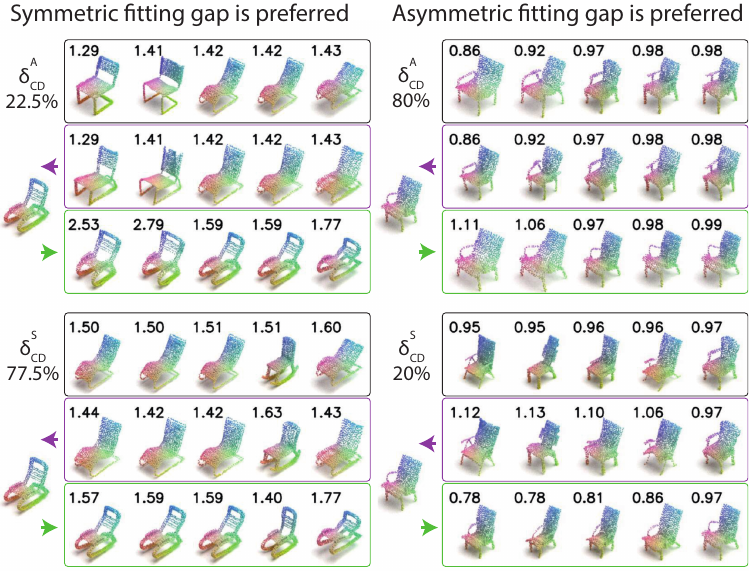}
   \vspace{-0.7cm}
   \caption{Examples of shapes for which asymmetric or symmetric fitting gaps are preferred (\cref{sec:study_ab}). 
    The shapes in the black rectangles show the top-5 ranked 3D shapes according to the asymmetric $\delta^A_{CD}$ or symmetric $\delta^S_{CD}$ fitting gaps.
    The shapes in the purple and green rectangles are the deformation results and their respective fitting gap values. 
    The arrows indicate the deformation direction. 
    Thus, the shapes in the purple rectangles are the gallery shapes deformed towards the query shape.
   }
   \label{fig:study_one_both}
   \vspace{-0.4cm}
\end{figure}

\begin{figure*}[ht]
    \centering
\includegraphics[width=\linewidth]{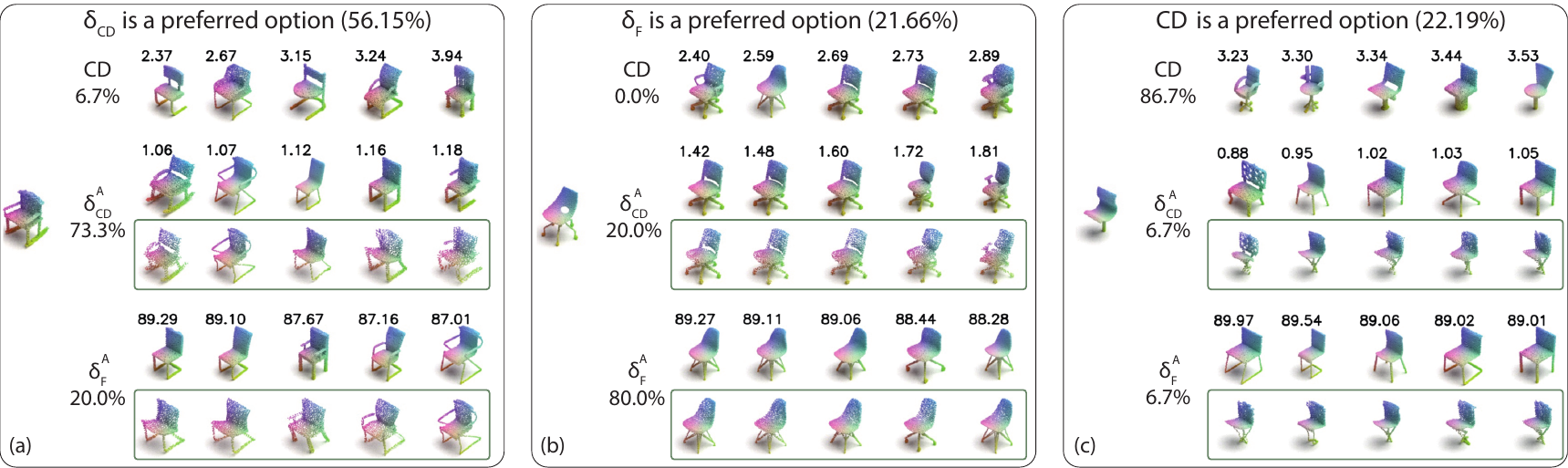}
    \vspace{-0.6cm}
    \caption{\emph{Human preferences on retrieval results according to different measures, excluding the groundtruth.} The Chamfer Distance ($CD$),  $\delta^{A}_{CD}$ (\cref{eq:fiting_gap}) and $\delta^{A}_F$  (\cref{eq:fiting_gap_F}) are selected by participants as the best measure reflecting shapes structure similarity for $16\%$ (4 shapes), $72\%$ (18 shapes) and $12\%$ (3 shapes) of shapes considered in the study, respectively. 
    The values below each measure symbol show the percentage of participants who selected a particular measure for a given shape as the best. 
    \new{The images in green rectangles show the deformations of the top-$5$ ranked shapes towards the query shape. 
    The numbers above each shape image are the values of the evaluated measure. 
    }
    }
    \label{fig:study}
    \vspace{-0.2cm}
\end{figure*}

\begin{figure}[t]
  \centering
  \includegraphics[width=1.0\linewidth]{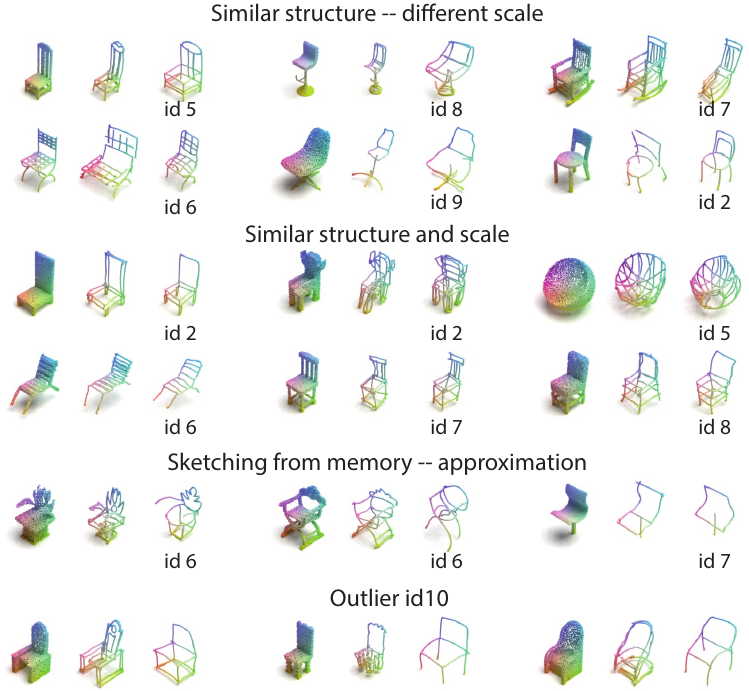}
   \vspace{-0.7cm}
  \caption{\emph{FVRS-M} vs.~\emph{FVRS}: In each triplet of images, the 1st image is a reference shape, the 2nd is a sketch from \emph{FVRS}\cite{luo2021FineGrainedVR}, and the 3rd is a sketch from our new dataset \emph{FVRS-M}. ``id'' denotes each participant's unique identifier. 
  The 1st row shows examples where the participants sketching from the visible reference preserved the overall proportions better, while the 2nd row shows cases where the participants sketching from memory reproduced the overall proportions better. 
  \emph{Large difference in proportions in these sketches further support our goal of designing a structure-aware retrieval method.} 
  The 3rd and 4th rows show examples in which participants in both sketching scenarios accurately reproduced structure and shape dimensions. 
  The 5th row shows that when sketching from memory, participants draw complex details approximately. 
  Finally, one participant drew too quick and non-detailed sketches (row 6).}
  \label{fig:FVRS-M}
\end{figure}

\section{Perceptual study}
In this section, we study the appropriateness of the fitting gap as an evaluation criterion of the structural similarities between 3D shapes, as judged by a human observer.
First, we compare the asymmetric and symmetric fitting gaps.
Second, we evaluate the asymmetric fitting gap against the Chamfer distance and F-score. 

\subsection{Symmetric vs. asymmetric fitting gap}
\label{sec:study_ab}

\paragraph{Data preparation} 
We select a subset of $24$ diverse shapes from the ShapeNet chair category, on which the top-5 retrieved shapes for two versions of the fitting gap differ visually strongly from each other (some  examples are shown in \cref{fig:study_one_both}).
Then, for each shape, we perform a nearest neighbor retrieval among the 5,274 shapes.
We represent shapes as point clouds by uniformly sampling 1,024 points.

\paragraph{Task} 
We showed to $39$ participants the top 5 ranked retrieval results according to two versions of the fitting gap. 
We asked the participants to select the retrieval results that \emph{have the most similar structure to the reference}, as judged from left to right. 

\paragraph{Results} 
For all shapes and all participants, the asymmetric fitting gap was chosen $49.50\%$ of the time, while the symmetric fitting gap was $50.50\%$ of the time. 
While on average there is no clear preference for one over the other, we observe that the best measure depends on a shape, as shown in \cref{fig:study_one_both}: on different shapes, one measure is chosen over another with a clear gap.  
\new{Which fitting gap is preferred is driven by how easy one or another deformation direction is.
Thus, the example on the left shows the shape for which computing deformation in both direction results in more reliable ranking. 
Meanwhile, for the example on the right the ranked results are dominated by the deformation direction from the query to the gallery shapes.} 

\subsection{Fitting gap vs. Chamfer distance}
\label{sec:study}
In this experiment, we select a subset of $25$ diverse shapes from the ShapeNet chair category. 
Then, for each shape, we perform a nearest neighbor retrieval among 5,274 shapes according to one of the 3 measures described below. 
The task in this study is the same as the one described in the previous section.

We evaluate the Chamfer distance against the two variants of the asymmetric fitting gap.
\new{We choose the asymmetric fitting gap because it requires less computation and performs similarly to the symmetric fitting gap on average.}
As the first variant of the fitting-gap-based measures, we consider an asymmetric fitting gap, where the distance after the deformation is computed via Chamfer distance (\cref{eq:fiting_gap}). 
We also evaluate the asymmetric fitting gap, based on the F-score with the threshold values set to $0.01$:
\begin{equation}
\delta^A_F(T_S^{p},T_S^{n}) \coloneqq \text{F-score}(\mathcal{D}(T_S^{n}; T_S^{p}),T_S^{p})
\label{eq:fiting_gap_F}
\end{equation}
It was observed that the Chamfer distance is sensitive to geometric layout of outliers, and F-score was proposed as an additional metric \cite{tatarchenko2019single}. 
The ShapeNet 3D shapes are normalized to fit a unit bounding box.


\paragraph{Results} 
We collect responses from $15$ participants.
\cref{fig:study} shows representative distributions of human preferences. 
On average the Chamfer distance $CD$ is preferred $22.19\%$ of the time, $\delta^{A}_{CD}$ (\cref{eq:fiting_gap}) is preferred $56.15\%$ of the time, and $\delta^{A}_{F}$ (\cref{eq:fiting_gap_F}) is preferred $21.66\%$ of the time.
We observe a good agreement between the participants: the standard deviations in choosing $CD$, $\delta^A_{CD}$ and $\delta^A_F$ comprise $6.85$, $7.34$ and $7.36$ respectively.
Yet, there is a larger spread in preferences among the shapes (\cref{fig:study}): the standard deviations in choosing $CD$, $\delta^{A}_{CD}$ and $\delta^{A}_F$ comprise $22.29$, $23.48$ and $19.26$ respectively.
\new{In \cref{fig:study}(c) it can be seen that the Chamfer distance is more likely to be selected when it is too `easy' to warp any shape to the source groundtruth shape, and the fitting gap is not informative. However, such cases are not very frequent.}
\new{It also can be seen that the F-score is not very good in capturing the differences between shapes due to small details (\cref{fig:study}(a)). 
Even though in \cref{fig:study}(b) $\delta^A_F$ is preferred by the majority of participants, it can be seen that the fitting gap with the Chamfer distance allows us to retrieve shapes with closer form of the chairs' legs, and containing a hole in the back similar to the ground-truth. }

\section{Experiments and Analysis}

\subsection{Datasets}
For training and testing we exploit the dataset of 3D VR sketches by Luo \etal \cite{luo2021FineGrainedVR}. 
This dataset consists of 1,005 unique 3D sketch and shape pairs from the ShapeNet chair category, created by 50 participants. 
VR sketches in this dataset were collected by displaying the reference shape in the area separate from the sketching area. 
We refer to this dataset as \emph{FVRS} -- a dataset of Freehand VR Sketches. 

We use the same split for testing, training and validation sets as was proposed in \cite{luo2021FineGrainedVR}, comprising $202$, $702$ and $101$ sketches, respectively.
During testing we perform retrieval from 5,794 shapes unseen during training and validation.
The test set contains all the sketches of 5 randomly selected participants (50 sketches), and a subset of sketches of each of the remaining 45 participants (152 sketches). 
The 45 participants' sketches are split proportionally across the training, validation and test sets.
We represent our sketches and shapes by uniformly sampling 1,024 points.

\new{As in our envisioned application scenario a human is sketching from memory or imagination, we collected a small set of sketches from memory to (1) test the performance on such data and (2) being able to analyze the differences with previously collected data from observation.}
\old{To further test how representative these \emph{FVRS} sketches are of those that people can draw from memory, we collect an additional dataset that we use for testing, where the participants first study the reference and then sketch it from memory.
The participants were asked to sketch the features of the reference shape that discriminate it from others.
And they could only sketch when the reference shape was invisible, ensuring that they needed to memorize the reference.}
\new{The participants could only sketch when the reference shape was invisible, ensuring that they needed to memorize the reference. They were instructed to sketch the features of the reference shape that discriminate it from others.}
They could go back to the reference at any time, as the task of memorizing an arbitrary shape is somewhat artificial. This allows to better approximate the scenario where people draw some concrete shapes from memory.
For a fair evaluation, we collect additional sketches \emph{for the same $202$ shapes  as the 202 sketches in the test set of the FVRS datasset}. We hire $10$ participants, where $8$ sketch $20$ shapes each and $2$ sketch $21$ shapes. We refer to this new dataset as \emph{FVRS-M} -- a dataset of Freehand VR Sketches by sketching from Memory. 
The analysis and comparison of sketches from \emph{FVRS-M} and \emph{FVRS} is provided in \cref{fig:FVRS-M} and its caption.

\subsection{Evaluation criteria}
We evaluate the accuracy of the retrieval using 4 different metrics. 
In all cases, our sketches, source shapes, deformed source and target shapes are represented as point clouds, and points-to-points distances are used for the evaluation.
First, we evaluate the results in terms of Top-$k$ accuracy -- the standard fine-grained retrieval metric. 
Top-$k$ accuracy computes the percentage of queries for which the ground-truth shape is among the top $k$ ranked retrieval results. 
In addition, we compute 3 additional criteria allowing us to make more informed conclusions on how similar the shapes are to the ground-truth among the top $k$ ranked retrieval results. 
First, we evaluate the similarity in terms of the average Chamfer distance (\emph{Avg CD}) between each of the shapes in the top-$k$ ranked results and the ground-truth.
Then, we evaluate the structural similarity (\emph{Avg $\delta^{A/S}_{CD}$}) based on the average asymmetric/symmetric fitting gap $\delta^{A/S}_{CD}$ between each of the shapes in the top-$k$ ranked results and the ground-truth.


\begin{table*}[ht]
\centering
\caption{Quantitative comparison (\cref{sec:results}) of the proposed method against state-of-the-art \cite{luo2021FineGrainedVR}. 
The usage of data augmentation with sketch and shape distortions is denoted by $*$.
The best result in each column is highlighted in bold, and the second best result is underlined.
\emph{Acc} denotes the top-$k$ retrieval accuracy -- the percentage of queries that contain a ground-truth among its top $k$ ranked retrieval results.
\emph{Avg $X$} denotes the average distance between the ground-truth 3D shape and the ranked top-$k$ retrieval results according to the criterion $X$:  $CD$ is the Chamfer distance, $\delta^{A/S}$ is the asymmetric/symmetric fitting gap,
$\delta_{CD}^{A/S}$ is defined in \cref{eq:fiting_gap} and  \cref{eq:fiting_gap_s},
for the $\delta_{F}^{A/S}$ the distance between the shapes after deformation is computed as an F-score with the threshold set to 0.01.
%
\emph{Avg $X$} values are multiplied by $1e2$
}
\vspace{-0.2cm}
\label{tab:results}
\begin{subtable}{\linewidth}
 \caption{On the sketches from the \emph{FVRS} dataset \cite{luo2021FineGrainedVR}.}
 \label{tab:results_a}
 \small{

\resizebox{\textwidth}{!}{%
\begin{tabular}{l|rrr|
r rr|rrr|rrr|rrr|rrr}
                                                     & \multicolumn{3}{c|}{Acc $\uparrow$}                                           & \multicolumn{3}{c|}{Avg $CD(T_S^{p},T_S^{n}) \downarrow$}                     & \multicolumn{3}{c|}{Avg $\delta^A_{CD}(T_S^{p},T_S^{n}) \downarrow$}          & \multicolumn{3}{c|}{Avg $\delta^S_{CD}(T_S^{p},T_S^{n}) \downarrow$}          & \multicolumn{3}{c|}{Avg $\delta^A_{F}(T_S^{p},T_S^{n}) \uparrow$}             & \multicolumn{3}{c}{Avg $\delta^S_{F}(T_S^{p},T_S^{n}) \uparrow$}             \\ \hline
Model                                                & \multicolumn{1}{c}{t-1} & \multicolumn{1}{c}{t-5} & \multicolumn{1}{c|}{t-10} & \multicolumn{1}{c}{t-1} & \multicolumn{1}{c}{t-5} & \multicolumn{1}{c|}{t-10} & \multicolumn{1}{c}{t-1} & \multicolumn{1}{c}{t-5} & \multicolumn{1}{c|}{t-10} & \multicolumn{1}{c}{t-1} & \multicolumn{1}{c}{t-5} & \multicolumn{1}{c|}{t-10} & \multicolumn{1}{c}{t-1} & \multicolumn{1}{c}{t-5} & \multicolumn{1}{c|}{t-10} & \multicolumn{1}{c}{t-1} & \multicolumn{1}{c}{t-5} & \multicolumn{1}{c}{t-10} \\ \hline
\multicolumn{1}{l|}{\cite{luo2021FineGrainedVR}}     & 24.55          & 43.76          & 52.57          & 3.93          & 4.88          & 5.17          & 1.27          & 1.59          & 1.69          & 1.23          & 1.56          & 1.67          & 85.29          & 81.44          & 80.18          & 85.69          & 81.67          & 80.35          \\
\multicolumn{1}{l|}{\cite{luo2021FineGrainedVR}$^*$}  & 24.95          & 44.46          & 53.37          & 3.50          & 4.64          & 4.99          & 1.15          & 1.51          & 1.63          & 1.12          & 1.50          & 1.62          & 86.58          & 82.36          & 80.94          & 86.91          & 82.50          & 81.00          \\\hline
CD                                               & 23.47          & 41.49          & 49.01          & 3.18          & {\ul 3.98}    & {\ul 4.22}    & 1.10          & 1.43          & 1.52          & 1.07          & 1.40          & 1.50          & 86.92          & 82.95          & 81.84          & 87.22          & 83.19          & 82.02          \\
CD$^*$                                           & 24.65          & 43.86          & 53.27          & \textbf{2.89} & \textbf{3.95} & \textbf{4.21} & 1.09          & 1.40          & {\ul 1.49}    & {\ul 1.06}    & {\ul 1.37}    & {\ul 1.47}    & 87.23          & 83.50          & 82.25          & 87.51          & 83.64          & 82.37          \\\hline
$\delta^{A}_{CD}$                                & 24.55          & 43.86          & 52.67          & 3.23          & 4.16          & 4.49          & 1.10          & 1.40          & 1.50          & 1.08          & 1.39          & 1.48          & 86.96          & 83.26          & 82.09          & 87.27          & 83.43          & 82.23          \\
${\delta^{A}_{CD}}^*$                            & \textbf{27.23} & \textbf{47.13} & \textbf{55.25} & 3.05          & 4.22          & 4.54          & \textbf{1.05} & \textbf{1.38} & {\ul 1.49}    & \textbf{1.03} & \textbf{1.36} & {\ul 1.47}    & {\ul 87.48}    & {\ul 83.57}    & {\ul 82.32}    & {\ul 87.73}    & {\ul 83.73}    & {\ul 82.45}    \\\hline
$\delta^{S}_{CD}$                                & 25.45          & 43.96          & 53.37          & 3.07          & 4.13          & 4.38          & 1.09          & 1.41          & 1.50          & {\ul 1.06}    & 1.38          & {\ul 1.47}    & 87.18          & 83.21          & 82.14          & 87.59          & 83.49          & 82.37          \\
${\delta^{S}_{CD}}^*$                            & {\ul 26.53}    & {\ul 46.14}    & {\ul 55.15}    & {\ul 2.93}    & 4.13          & 4.49          & {\ul 1.06}    & {\ul 1.39}    & \textbf{1.48} & \textbf{1.03} & \textbf{1.36} & \textbf{1.45} & \textbf{87.63} & \textbf{83.71} & \textbf{82.39} & \textbf{88.01} & \textbf{83.95} & \textbf{82.57}
\\ \hline
\end{tabular}%
}
 }
 \end{subtable}
\begin{subtable}{\linewidth}
\centering
 \caption{On the sketches from the \emph{FVRS-M} dataset.}
 \vspace{-0.8pt}
 \label{tab:results_b}
\footnotesize{
 \begin{tabular}{l|rrr|rrr|rrr}
\multicolumn{1}{c|}{}                         & \multicolumn{3}{c|}{Acc $\uparrow$}                                                 & \multicolumn{3}{c|}{Avg $CD(T_S^{p},T_S^{n}) \downarrow$}                           & \multicolumn{3}{c}{Avg $\delta^A_{CD}(T_S^{p},T_S^{n}) \downarrow$}                            \\
\multicolumn{1}{c|}{\multirow{-2}{*}{Method}} & \multicolumn{1}{c}{top-1} & \multicolumn{1}{c}{top-5} & \multicolumn{1}{c|}{top-10} & \multicolumn{1}{c}{top-1} & \multicolumn{1}{c}{top-5} & \multicolumn{1}{c|}{top-10} & \multicolumn{1}{c}{top-1}             & \multicolumn{1}{c}{top-5} & \multicolumn{1}{c}{top-10} \\ \hline
\cite{luo2021FineGrainedVR}       & 15.94          & 34.55          & 43.76          & 4.63         & 5.27          & 5.55          & 1.43          & 1.64          & 1.74          \\
{\cite{luo2021FineGrainedVR}$^*$} & 17.92          & {\ul 36.63}    & 44.36          & 4.18         & 5.04          & 5.39          & 1.32          & 1.6           & 1.71          \\ \hline
$\delta^A_{CD}$                   & 17.43          & 35.35          & 43.66          & {\ul 3.81}   & {\ul 4.58}    & {\ul 4.87}    & {\ul 1.24}    & {\ul 1.46}    & 1.55          \\
$\delta^A_{CD}$$^*$               & \textbf{19.41} & \textbf{38.42} & \textbf{47.33} & 3.91         & 4.6           & {\ul 4.87}    & \textbf{1.22} & \textbf{1.43} & \textbf{1.51} \\ \hline
$\delta^S_{CD}$                   & 16.63          & 34.36          & 43.17          & 3.91         & 4.6           & {\ul 4.87}    & 1.26          & 1.49          & 1.55          \\
$\delta^S_{CD}$$^*$               & {\ul 18.51}    & 36.53          & {\ul 45.45}    & \textbf{3.8} & \textbf{4.57} & \textbf{4.83} & \textbf{1.22} & \textbf{1.43} & {\ul 1.52}   
\\
\hline

\end{tabular}
}

\end{subtable}

\end{table*}

\subsection{Results}
\label{sec:results}


All models are evaluated by training for 500 epochs.
We evaluate using cross-validation. 
We keep the test set fixed and partition the rest of the data (803 sketch-shape pairs) into 5 different subsets. 
For each of the 45 sketchers, we split their sketches proportionally, randomly, between the validation and training sets.
We select 101 sketches as the validation set and 702 sketches as the training set. 
The results in the paper are obtained by the last checkpoint. Results chosen by other metrics on validation set can be found in supplemental materials. 

\begin{figure*}[ht]
  \centering
\includegraphics[width=1.0\linewidth]{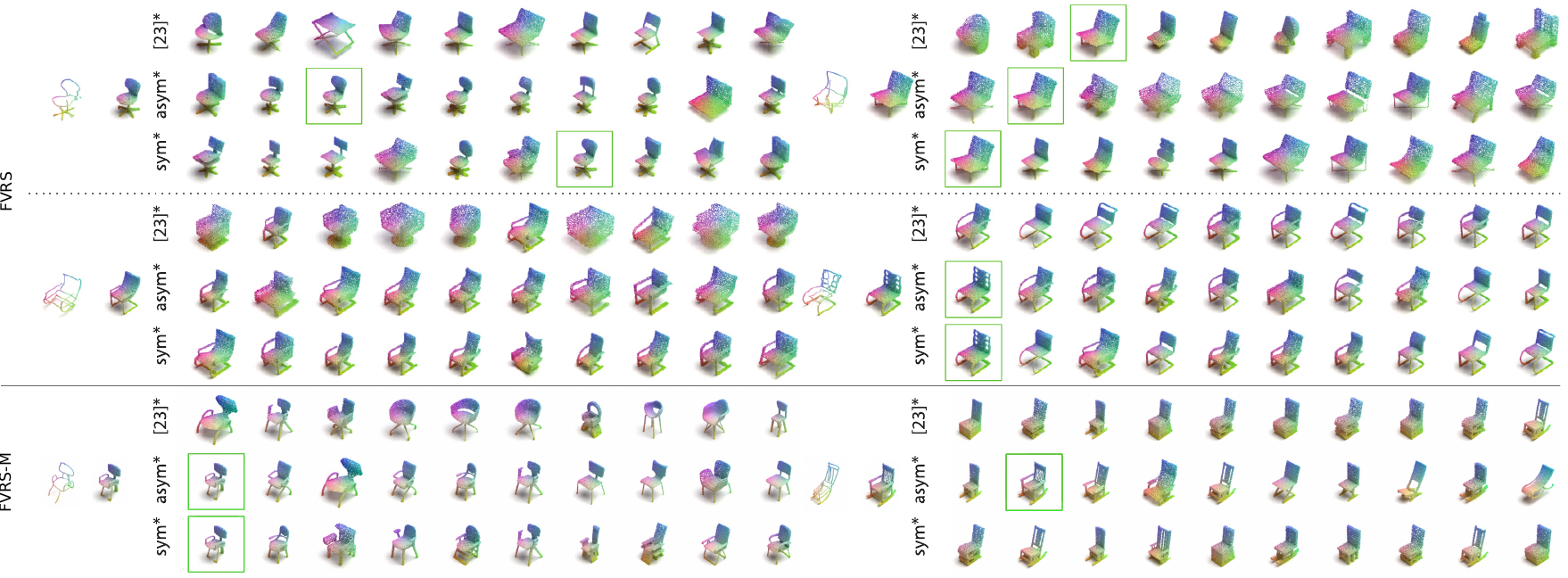}
   \vspace{-0.7cm}
   \caption{Qualitative evaluation: The lines in each sub-figure show top-10 ranked retrieval results when data augmentation is used for \cite{luo2021FineGrainedVR}, $\delta^A_{CD}$ (asym*) and $\delta^S_{CD}$ (sym*) models. 
   The green square denotes the ground-truth matching shape. 
   }
   \label{fig:results}
\vspace{-1.2pt}
\end{figure*}

\subsubsection{Adaptive vs. fixed margin}


First, we compare our model against the state-of-the-art, comprising training with a triplet loss based model with a constant margin value \cite{luo2021FineGrainedVR} with and without data augmentation (\cref{tab:results}). 
As data augmentation, we apply global sketch/shape distortions as was proposed by Luo \etal ~\cite{luo2021FineGrainedVR}.

We tested three constant margin values in the range $[0.3, 0.9]$ and found the value of $0.6$ to give the best top-$k$ retrieval accuracy and $0.3$ to perform better according to other criteria. 
When the data augmentation is used, the margin value of $0.6$ consistently outperforms other constant margin values settings. Please see the supplemental for the detailed comparison. 
The results in \cref{tab:results} are computed with the margin value of $0.6$. 

We set our method hyper-parameters $\alpha$ and $\beta$ in \cref{eq:margin} to $0.3$ and $1.2$, respectively.
The ratio on the right in \cref{eq:margin} takes values in the range $(0,1]$, so the margin value is in the range $(0.3,1.2]$. 
Our ablation study in the supplemental shows that our method is relatively robust to small variations in the values of these parameters, and consistently outperforms the fixed margin setting. 

%
\cref{tab:results} shows that the proposed adaptive margin scheme results in better performance according to all measures evaluating the average similarity (\emph{Avg $X$}) for the top-$1,5,10$ ranked retrieval results -- retrieving the 3D shapes with better structural similarities to the sketch query's ground-truth 3D shape (\cref{tab:results}).
Moreover, the fitting gap based baselines (the last 4 rows in \cref{tab:results_a}, \cref{tab:results_b}) also have higher top-$k$ retrieval accuracy.

Importantly, we observe that for the models with the fixed margin value, the average fitting gap value (\emph{Avg $\delta^{A/S}_{CD}$}) of the top-5 ranked retrieved results on validation set starts to increase after roughly the 30th epoch, and only the fitting gap value for the top-1 ranked retrieved results decreases. 
This may not be desirable if a user is interested in finding a set of similar shapes, rather than only one. 


\begin{table*}[ht]
\centering
\caption{Quantitative comparison of the proposed method against state-of-the-art on lamp and airplane shapes. Please see \cref{tab:results} for the notations.}
\vspace{-0.4cm}
\label{tab:other_category}
\resizebox{\textwidth}{!}{%
\begin{tabular}{l|l|crr|crr|crr|crr|crr|crr}
                       &       & \multicolumn{3}{c|}{Acc $\uparrow$}                                                      & \multicolumn{3}{c|}{Avg $CD(T_S^{p},T_S^{n}) \downarrow$}                               & \multicolumn{3}{c|}{Avg $\delta^A_{CD}(T_S^{p},T_S^{n}) \downarrow$}                    & \multicolumn{3}{c|}{Avg $\delta^S_{CD}(T_S^{p},T_S^{n}) \downarrow$}                    & \multicolumn{3}{c|}{Avg $\delta^A_{F}(T_S^{p},T_S^{n}) \uparrow$}                        & \multicolumn{3}{c}{Avg $\delta^S_{F}(T_S^{p},T_S^{n}) \uparrow$}                        \\ \hline
\multicolumn{1}{c|}{Category}  & \multicolumn{1}{c|}{Model}    & t-1                                & \multicolumn{1}{c}{t-5} & \multicolumn{1}{c|}{t-10} & t-1                               & \multicolumn{1}{c}{t-5} & \multicolumn{1}{c|}{t-10} & t-1                               & \multicolumn{1}{c}{t-5} & \multicolumn{1}{c|}{t-10} & t-1                               & \multicolumn{1}{c}{t-5} & \multicolumn{1}{c|}{t-10} & t-1                                & \multicolumn{1}{c}{t-5} & \multicolumn{1}{c|}{t-10} & t-1                                & \multicolumn{1}{c}{t-5} & \multicolumn{1}{c}{t-10} \\ \hline
\multicolumn{1}{l|}{\multirow{3}{*}{lamp}}     & {\cite{luo2021FineGrainedVR}$^*$} & 63.87                   & 83.23                   & 86.45                    & 1.20                    & 6.37                    & 8.82                     & 0.39                    & 1.54                    & 1.98                     & 0.43                    & 1.86                    & {\ul 2.43}               & 95.38                   & 82.57                   & 78.09                    & 94.96                   & 81.55                   & 76.51                    \\
\multicolumn{1}{l|}{}                          & $\delta^A_{CD}$$^*$               & \textbf{67.74}          & \textbf{87.10}          & \textbf{92.26}           & {\ul 1.11}              & {\ul 4.00}              & {\ul 5.08}               & \textbf{0.33}           & {\ul 1.24}              & \textbf{1.45}            & \textbf{0.32}           & \textbf{1.35}           & \textbf{1.64}            & \textbf{96.11}          & {\ul 85.29}             & {\ul 82.67}              & \textbf{95.99}          & {\ul 85.02}             & \textbf{82.06}           \\
\multicolumn{1}{l|}{}                          & $\delta^S_{CD}$$^*$               & {\ul 65.81}             & {\ul 85.16}             & {\ul 91.61}              & \textbf{0.98}           & \textbf{3.78}           & \textbf{4.98}            & {\ul 0.34}              & \textbf{1.21}           & {\ul 1.46}               & {\ul 0.34}              & {\ul 1.31}              & \textbf{1.64}            & {\ul 95.95}             & \textbf{85.71}          & \textbf{82.75}           & {\ul 95.96}             & \textbf{85.54}          & {\ul 82.24}              \\ \hline
\multicolumn{1}{l|}{\multirow{3}{*}{airplane}} & {\cite{luo2021FineGrainedVR}$^*$} & 81.33                   & 92.00                   & 93.33                    & 0.24                    & 1.34                    & 1.69                     & {\ul 0.16}              & 0.59                    & 0.68                     & {\ul 0.16}              & 0.62                    & 0.72                     & 99.27                   & 93.58                   & 92.40                    & 99.24                   & 93.29                   & 92.03                    \\
\multicolumn{1}{l|}{}                          & $\delta^A_{CD}$$^*$               & {\ul 82.22}             & {\ul 93.78}             & {\ul 95.56}              & \textbf{0.20}           & {\ul 1.11}              & {\ul 1.40}               & \textbf{0.14}           & {\ul 0.52}              & {\ul 0.61}               & \textbf{0.15}           & {\ul 0.53}              & {\ul 0.63}               & \textbf{99.44}          & {\ul 94.69}             & {\ul 93.48}              & \textbf{99.39}          & {\ul 94.54}             & {\ul 93.25}              \\
\multicolumn{1}{l|}{}                          & $\delta^S_{CD}$$^*$               & \textbf{82.67}          & \textbf{96.00}          & \textbf{96.00}           & {\ul 0.23}              & \textbf{1.08}           & \textbf{1.33}            & \textbf{0.14}           & \textbf{0.49}           & \textbf{0.57}            & \textbf{0.15}           & \textbf{0.51}           & \textbf{0.59}            & {\ul 99.38}             & \textbf{94.97}          & \textbf{93.93}           & {\ul 99.33}             & \textbf{94.83}          & \textbf{93.75}          

\\ \hline
\end{tabular}%
}
\end{table*}

\begin{figure*}[ht]
\vspace{-0.2cm}
  \centering
\includegraphics[width=1.0\linewidth]{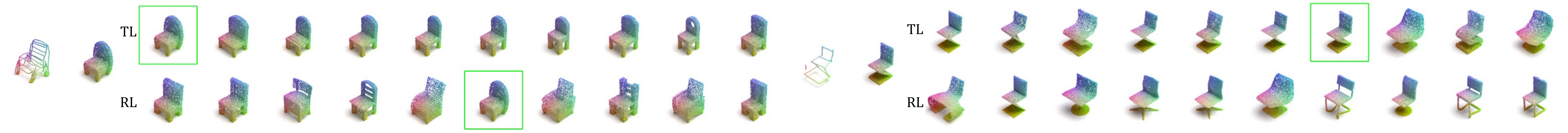}
  \vspace{-0.5cm}
   \caption{Qualitative comparison of the regression loss (RL) \cite{uy2020deformation} and the triplet loss (TL) with an adaptive margin (\cref{eq:margin,eq:TL}). The lines in each sub-figure show top-10 ranked retrieval results for (1) TL: $\delta^{A}_{CD}$ with triplet loss  , and (2) RL: $\delta^{A}_{CD}$ with regression loss. The green square denotes the ground-truth matching shape.
   }
   \label{fig:results_TL_vs_RL}
\end{figure*}

\begin{table*}[ht!]
\centering
\caption{Comparison of the regression loss (RL) \cite{uy2020deformation} and the triplet loss (TL) with an adaptive margin (\cref{eq:margin,eq:TL}).
Please see \cref{tab:results} for the notations.
}
\label{tab:regression_loss}
\vspace{-0.4cm}
\resizebox{\textwidth}{!}{%
\begin{tabular}{l|crr|crr|crr|crr|crr|crr}
 \multicolumn{19}{c}{On the sketches from the \emph{FVRS} dataset \cite{luo2021FineGrainedVR}.}  \\ \hline
 
                              & \multicolumn{3}{c|}{Acc $\uparrow$}                                                      & \multicolumn{3}{c|}{Avg $CD(T_S^{p},T_S^{n}) \downarrow$}                               & \multicolumn{3}{c|}{Avg $\delta^A_{CD}(T_S^{p},T_S^{n}) \downarrow$}                    & \multicolumn{3}{c|}{Avg $\delta^S_{CD}(T_S^{p},T_S^{n}) \downarrow$}                    & \multicolumn{3}{c|}{Avg $\delta^A_{F}(T_S^{p},T_S^{n}) \uparrow$}                        & \multicolumn{3}{c}
                              {Avg $\delta^S_{F}(T_S^{p},T_S^{n}) \uparrow$}                        \\ \hline
\multicolumn{1}{c|}{Model}    & t-1                                & \multicolumn{1}{c}{t-5} & \multicolumn{1}{c|}{t-10} & t-1                               & \multicolumn{1}{c}{t-5} & \multicolumn{1}{c|}{t-10} & t-1                               & \multicolumn{1}{c}{t-5} & \multicolumn{1}{c|}{t-10} & t-1                               & \multicolumn{1}{c}{t-5} & \multicolumn{1}{c|}{t-10} & t-1                                & \multicolumn{1}{c}{t-5} & \multicolumn{1}{c|}{t-10} & t-1                                & \multicolumn{1}{c}{t-5} & \multicolumn{1}{c}{t-10} \\ \hline
\textbf{$\delta^{A}_{CD}$ TL} & \textbf{24.55} & \textbf{43.86} & \textbf{52.67} & \textbf{3.23} & 4.16          & 4.49          & \textbf{1.10} & 1.40          & 1.50          & \textbf{1.08} & 1.39          & 1.48          & \textbf{86.96} & 83.26          & 82.09          & \textbf{87.27} & \textbf{83.43} & 82.23          \\
$\delta^{A}_{CD}$ RL & 15.35          & 31.98          & 39.11          & 3.62          & \textbf{4.12} & \textbf{4.28} & 1.22          & \textbf{1.35} & \textbf{1.41} & 1.23          & \textbf{1.38} & \textbf{1.44} & 84.85          & \textbf{83.42} & \textbf{82.81} & 84.88          & 83.31          & \textbf{82.68}
\\ \hline
\multicolumn{19}{c}{On the sketches from the \emph{FVRS-M} dataset.} 
\\ \hline
                              & \multicolumn{3}{c|}{Acc $\uparrow$}                                                      & \multicolumn{3}{c|}{Avg $CD(T_S^{p},T_S^{n}) \downarrow$}                               & \multicolumn{3}{c|}{Avg $\delta^A_{CD}(T_S^{p},T_S^{n}) \downarrow$}                    & \multicolumn{3}{c|}{Avg $\delta^S_{CD}(T_S^{p},T_S^{n}) \downarrow$}                    & \multicolumn{3}{c|}{Avg $\delta^A_{F}(T_S^{p},T_S^{n}) \uparrow$}                        & \multicolumn{3}{c}
                              {Avg $\delta^S_{F}(T_S^{p},T_S^{n}) \uparrow$}                        \\ \hline
\multicolumn{1}{c|}{Model}    & t-1                                & \multicolumn{1}{c}{t-5} & \multicolumn{1}{c|}{t-10} & t-1                               & \multicolumn{1}{c}{t-5} & \multicolumn{1}{c|}{t-10} & t-1                               & \multicolumn{1}{c}{t-5} & \multicolumn{1}{c|}{t-10} & t-1                               & \multicolumn{1}{c}{t-5} & \multicolumn{1}{c|}{t-10} & t-1                                & \multicolumn{1}{c}{t-5} & \multicolumn{1}{c|}{t-10} & t-1                                & \multicolumn{1}{c}{t-5} & \multicolumn{1}{c}{t-10} \\ \hline
\textbf{$\delta^{A}_{CD}$ TL}                & \textbf{17.43} & \textbf{35.35} & \textbf{43.66} & \textbf{3.81} & 4.58          & 4.87         & \textbf{1.24} & 1.46          & 1.55          & \textbf{1.26} & \textbf{1.49} & 1.59          & \textbf{85.07} & \textbf{82.54} & 81.45          & \textbf{85.19} & \textbf{82.53} & 81.41          \\
$\delta^{A}_{CD}$ RL & 9.01           & 21.88          & 28.32          & 4.35          & \textbf{4.57} & \textbf{4.70} & 1.36          & \textbf{1.43} & \textbf{1.46} & 1.40           & \textbf{1.49} & \textbf{1.53} & 83.01          & 82.32          & \textbf{81.99} & 82.99          & 82.09          & \textbf{81.71}

\\ \hline
\end{tabular}%
}
\end{table*}

On the newly collected sketches from memory \emph{FVRS-M} we obtain lower retrieval accuracy on average (\cref{tab:results_b}) as compared to performance on sketches from \emph{FVRS} (\cref{tab:results_a}). 
Our adaptive margin scheme results in better performance according to all criteria: top-$k$ retrieval accuracy and measures evaluating the similarity of the top-$k$ retrieval results.


To demonstrate that our approach is not limited to a chair category, we generate synthetic 3D VR sketches for two additional categories: `lamp' and `airplane'. 
We rely on the method proposed in \cite{luo2020towards} and set the abstractness level to $1.0$, as it was demonstrated to match freehand sketches best. 
However, as we demonstrate in the supplemental their method still produces very dense sketches for the shapes from the `airplane' category, closely resembling the 3D shapes.
We show in \cref{tab:other_category} that our method with an adaptive margin outperforms constant margin setting in terms of top-$k$ accuracy and all distance metrics.



\subsubsection{Comparison with regression loss}

For 3D shape to 3D shape retrieval, Uy et al.~\cite{uy2020deformation} proposed a regression loss, to retrieve a 3D shape best matching a query 3D model after deformation.
Here, we adopt this loss to our problem, and use our definition of the asymmetric fitting gap as the distance between 3D shapes, and the $L_2$ distance as the distance in the feature space. Please see the supplemental for details.

\cref{tab:regression_loss} shows that, in certain cases, Regression Loss (RL) results in slightly better criteria, measuring the average similarity of the top-5/10 retrieval results to the ground-truth.  
However, RL results in significantly worse retrieval performance according to top-$k$ retrieval accuracy ($Acc$) and the criteria evaluating the average similarity of the top-1 retrieval results.
For additional evaluation and details please refer to the supplemental. 

RL encourages the distribution of distances in the latent space to approximate the distribution of distances between 3D shapes. 
Thus, many similar shapes are encoded close to each other in the latent space.
Meanwhile, triplet loss's primary goal is to ensure that the distance between similar shapes is smaller than the distance between dissimilar shapes.
Our formulation with the adaptive margin allows to achieve a desired trade off (\cref{fig:results_TL_vs_RL}).

\section{Conclusion and future work}
In this work, we propose an \emph{adaptive margin setting} allowing for \emph{structural similarities of shapes} to play an important role in ranking retrieval results.
We further show that, due to inaccuracies inherent to sketches, top-$k$ retrieval accuracy, the standard for retrieval evaluation, might be insufficient to drive out conclusions on retrieval performances. 
We deploy a recent off-the-shelf neural deformation operator to compute the fitting gap that measures the similarity between two 3D shapes under deformation. 
We conduct a user study that shows the superiority of this fitting gap over the Chamfer distance in measuring structural similarities. 
We deploy fitting gap as one of the criteria to measure retrieval results and as a distance metric for the adaptive margin scheme. 
We show that our results outperform state-of-the-art retrieval methods for sketch-based 3D shape retrieval.
Our perceptual studies show that both symmetric and asymmetric fitting gap can lead to sub-optimal rankings.
In the future, we would like to improve the robustness of the fitting gap driven retrieval by considering not only the distances after the deformation, but also how much the shapes need to be deformed.

\clearpage
{\small
\bibliographystyle{ieee_fullname}
\bibliography{camera_ready}
}

\end{document}